%% file: main.tex
\documentclass[runningheads]{llncs}
\usepackage[T1]{fontenc}
\usepackage{graphicx}
\usepackage{multirow}
\usepackage{array}
\usepackage{amsmath}
\usepackage{amsfonts}
\begin{document}
\title{Multi-Modal Adapter for Vision-Language Models}
%
%
\author{Dominykas Seputis\inst{1,2} \and
Serghei Mihailov\inst{1} \and
Soham Chatterjee\inst{1} \and Zehao Xiao\inst{1}}
\authorrunning{Seputis et al.}
%
\institute{University of Amsterdam, Amsterdam, Netherlands \\ \email{\{dominykas.seputis, serghei.mihailov, soham.chatterjee2 \}@student.uva.nl, z.xiao@uva.nl} \and
Vinted, Vilnius, Lithuania \\ \email{dominykas.seputis@vinted.com}}
\maketitle              
\begin{abstract}
Large pre-trained vision-language models, such as CLIP \cite{radford2021learning}, have demonstrated state-of-the-art performance across a wide range of image classification tasks, without requiring retraining. Few-shot CLIP is competitive with existing specialized architectures that were trained on the downstream tasks. Recent research demonstrates that the performance of CLIP can be further improved using lightweight adaptation approaches. 
However, previous methods adapt different modalities of the CLIP model individually, ignoring the interactions and relationships between visual and textual representations.
In this work, we propose Multi-Modal Adapter, an approach for Multi-Modal adaptation of CLIP. Specifically, we add a trainable Multi-Head Attention layer that combines text and image features to produce an additive adaptation of both. Multi-Modal Adapter demonstrates improved generalizability, based on its performance on unseen classes compared to existing adaptation methods. We perform additional ablations and investigations to validate and interpret the proposed approach.
\end{abstract}


\input{sections/1_intro}

\input{sections/2_related_work}

\input{sections/3_approach}

\input{sections/4_experiments_results}

\input{sections/5_conclusion}

\bibliographystyle{splncs04}
\bibliography{main}

\input{sections/appendix}
\end{document}

%% file: sections/1_intro.tex
\section{Introduction}
\label{sec:intro}

Large pre-trained vision-language models (VLMs) have demonstrated state-of-the-art capabilities by learning the joint embedding space of texts and images. 
Pre-training on vast data enables VLMs to learn diverse concepts, ensuring robust performance across tasks. Generalizability is crucial for applying pre-trained models across diverse tasks with minimal adaptation. A common approach to assess generalizability, which we use in this work, is evaluating the models in an \textit{$n$-class-$k$-shot} setting, where for a given task, the model is trained on $k$ samples from $n$ base classes, with unseen samples and classes during evaluation.

CLIP \cite{radford2021learning}, a prominent VLM, exceeds in zero-shot settings on new tasks. It utilizes contrastive learning to align text and image representations in a shared embedding space. 
For image classification tasks, CLIP predicts based on the similarity between text and image embeddings.

Recently, there has been growing research into adaptation strategies to improve CLIP's performance on downstream tasks. 
Prompt-based approaches, such as Context Optimization (CoOp) \cite{Zhou_2022} and  Conditional Context Optimization (CoCoOp) \cite{zhou2022conditional} adapt text encoder prompts, with CoCoOp also conditioning on image features.
On the other hand, there are adapter-based approaches. \cite{gao2021clipadapter} propose CLIP-Adapter, which fine-tunes two lightweight 2-layer MLPs on top of the image and text embeddings, respectively. 
Despite its simplicity, this approach has shown promising performance, comparable with prompt-based methods. 
However, these methods adapt image and text embeddings individually, which ignores the interaction between different modatwo-layer multi-layer perceptrons (MLPs) on top of the image and text embeddings, respectivelylities.
Moreover, the adaptation parameters are easy to overfit training data, leading to loss of the generalization ability for new tasks.

\begin{figure*}
\centering
\includegraphics[width=1\linewidth]{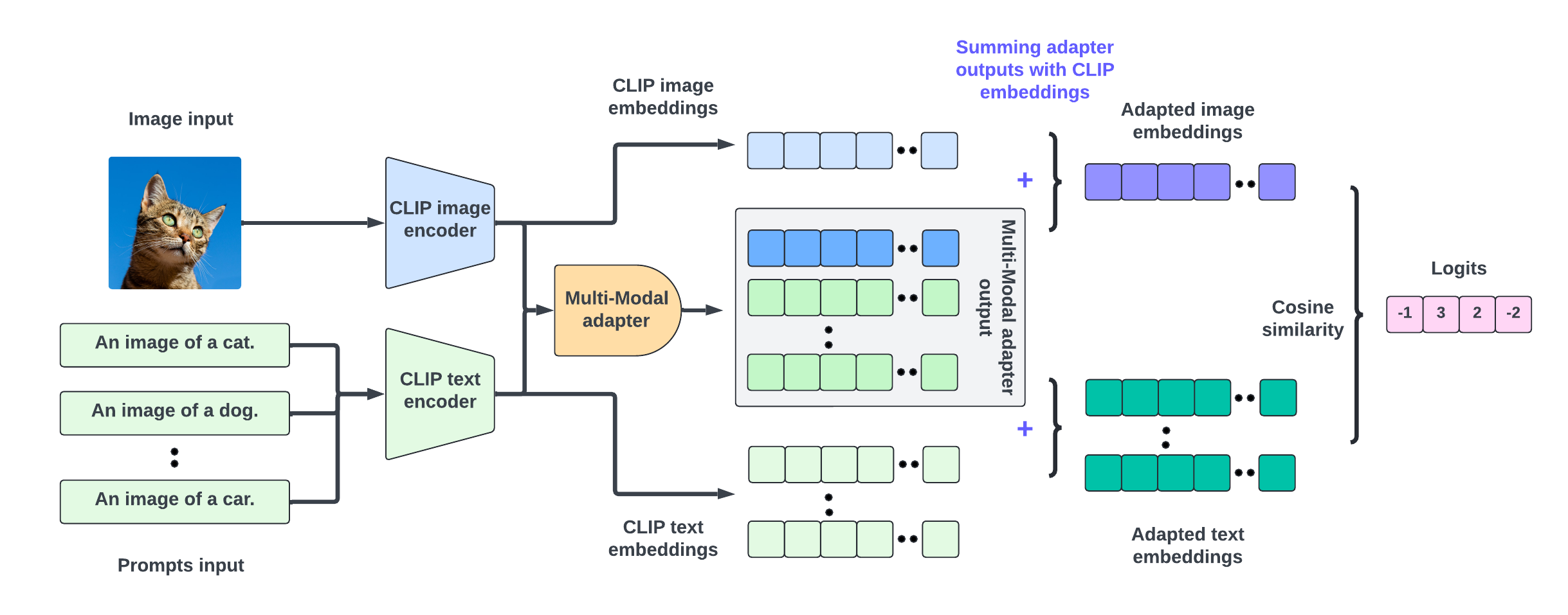}
\caption{Illustration of Multi-Modal adapter.}
\vspace{-10pt}
\label{fig:mm_adapter}
\end{figure*}

To address this limitation, we use an attention-based adapter, which we call Multi-Modal Adapter (Figure \ref{fig:mm_adapter}), jointly adapts the text and image features by integrating their information, leading to more comprehensive adaptations.
Moreover, adaptation is task-specific, utilizing the visual and textual information of each task effectively. This enables the model to more effectively avoid overfitting on training tasks, leading to better generalization ability on unseen tasks.
Our contribution can be summarized as follows:
\begin{itemize}
    \item We propose Multi-Modal Adapter that performs Multi-Modal information aggregation, allowing for joint adaptation of modalities.
    \item With task-specific fine-tuning, Multi-Modal Adapter achieves competitive results across a range of datasets in a n-class-k-shot setting. Multi-Modal Adapter improves generalizability, as demonstrated by its performance on unseen classes.
    \item Additionally, we perform ablation studies to validate the design and configuration of Multi-Modal Adapter.
\end{itemize}

The code is made available at \url{https://github.com/dqmis/clip-mma}.

%% file: sections/2_related_work.tex
\section{Related work}
\label{sec:related_work}
\subsection{Vision-Language Models}
Vision-language models have developed rapidly over the recent years. VLMs demonstrate advanced vision-language understanding (CLIP, ALIGN \cite{jia2021scaling}, VLMo \cite{bao2022vlmo}, ImageBind \cite{girdhar2023imagebind}) 
, text generation capabilities (GPT-4V \cite{openai2024gpt4}, Flamingo \cite{alayrac2022flamingo}, Frozen \cite{tsimpoukelli2021multimodal}), including multi-modal generation (Gemini \cite{geminiteam2024gemini}). The VLM approaches focus on aligning text and image data in a shared embedding space, to then be compared or treated as a unified sequence. Trained on vast amounts of data across vision and text, these models have the emergent ability of few-shot learning: understanding diverse concepts and displaying high performance without task-specific training.

Our primary focus is on CLIP, a dual-encoder model for vision-language understanding trained using contrastive loss. In the contrastive setting, models learn from $N$ image-text pairs by maximizing similarity for correct pairs and minimizing it for others. Combined with the Transformer architecture \cite{vaswani2017attention} and large-scale data, this approach is scalable and efficient for learning robust text and image representations in a shared space. Our Multi-Modal adapter approach is aimed at jointly improving the representations on downstream tasks. In this work, we validate our approach on CLIP, however it could also be applied to other architectures aligning Multi-Modal embeddings.

\subsection{Parameter-efficient fine-tuning (PEFT)}
The growing scale of pre-trained models has motivated the development of efficient fine-tuning techniques that utilize existing knowledge with minimal adaptations.

\textbf{Low-rank adaptation methods}, such as LoRA \cite {hu2021lora} and DoRA \cite{liu2024dora}, introduce trainable low-rank matrices for approximating weight updates during fine-tuning. In contrast, we introduce an additional adapter module that takes inputs from multiple modalities.

\textbf{Prompt-based methods}, such as prompt-tuning \cite{lester2021power} and prefix tuning \cite{li2021prefixtuning}, propose learning the input (or input prefix) as a series of trainable embedding, instead of fine-tuning parameters. In the context of CLIP prompt-tuning, Context Optimization (CoOp) and the subsequent Conditional Context Optimization (CoCoOp) are two notable approaches. CoOp learns task-specific context tokens appended to the text encoder input. Conditional Context Optimization (CoCoOp) builds upon CoOp by making the context tokens image-dependent, thus conditioning the text encoder on both the task and the specific image. These methods have proven to be effective for CLIP fine-tuning, however create additional overhead at inference, due to increasing input size and are in general sub-optimal compared to PEFT approaches. 

\textbf{Adapter-based methods} introduce trainable modules into pre-trained models \cite{houlsby2019parameterefficient,rebuffi2017learning}. CLIP-specific adapters, such as CLIP-Adapter \cite{gao2021clipadapter}, Tip-Adapter \cite{zhang2021tipadapter} and XMAdapter \cite{yang2024crossmodal}, modify outputs from image and text encoders to enhance embeddings. In CLIP-Adapter, a simple and representative architecture is used: text and image embeddings are adapted with 2-layer MLPs each. The resulting embeddings are computed as a weighted average of the original embedding and the adapted embedding. In contrast, our approach introduces a Multi-Modal Adapter to jointly adapt modalities, enabling greater interaction between text and image features, rather than relying solely on cosine similarity in the shared embedding space.

\textbf{Transformers for combining CLIP image and text features.} Our work is inspired by the Any-Shift Prompting approach \cite{xiao2024any}, in which a transformer combines test information from both image and text features as well as training prompts to generate a test-specific prompt. This method enhances prediction robustness across distribution shifts. 

%% file: sections/3_approach.tex
\section{Method}
\label{sec:approach}
\subsection{Preliminary}
\paragraph{CLIP.} CLIP \cite{radford2021learning} is a vision-language model designed for zero-shot learning. It comprises two encoders: a visual encoder, which can be a CNN like ResNet-50 \cite{he2015deep} or a ViT \cite{dosovitskiy2020image}, to map images into a low-dimensional embedding space, and a text encoder, built on a Transformer \cite{vaswani2017attention}, to generate text representations. CLIP predicts whether an image matches a textual description by comparing image features with the text ones using cosine similarity between the vectors. Unlike traditional classifiers that learn visual concepts from random vectors, CLIP's vision-language pre-training enables exploration of open-set visual concepts through a high-capacity text encoder, leading to a broader semantic space and more transferable representations for downstream tasks.

\paragraph{CLIP-Adapter.} \cite{gao2024clip} proposed an efficient method for adapting the CLIP model to new domains. The adapter appends a small number of additional learnable bottleneck linear layers with ReLu activation in between them to CLIP’s language and image encoders, while keeping the original CLIP backbone frozen. To mitigate overfitting and enhance the robustness of CLIP-Adapter, residual connections are utilized to dynamically blend the fine-tuned knowledge with the original knowledge from CLIP’s backbone.

\subsection{Multi-Modal Adapter}
We propose an iterative update to the CLIP-Adapter \cite{gao2024clip}, where instead of using linear transformations to adapt image and text embeddings individually, we employ a Multi-Modal (MM) adapter base on Multi-Head  Attention (MHA) network \cite{vaswani2017attention}. 
The network aggregates Multi-Modal information of both textual and visual data representations and adapts the Multi-Modal embeddings jointly. 
By integrating task-specific information across modalities, our adapter achieves specific adaptation for each task, which avoids overfitting training tasks.

Our approach comprises three distinct parts: an embedding downsampler, a masked Multi-Head attention network, and two linear layers with a non-linear activation function in between them.

\paragraph{Dimension downsampling.}
We use Multi-Head attention layers in our adapter to aggregate Multi-Modal information.
However, Multi-Head attention layers can introduce parameter-wise costly operations to the network, especially with large embedding dimensions, e.g., the original CLIP's embedding dimensions ($\text{C}_{\text{Emb}}$) $=$ 512. 
To lower the count of trainable parameters, we introduce a dimension downsampler on the CLIP embeddings before the attention layers. 
The downsampler consists of linear layer $D(\cdot)$ that reduce the embedding dimension. Finally, the input to the Multi-Head attention is passed through $D(\cdot)$, where $W_{D} \in \mathbb{R}^{\text{C}_{\text{Emb}} \times \text{C}_{\text{Emb}} / 4}$.

\begin{figure}
\centering
\includegraphics[width=0.55\linewidth]{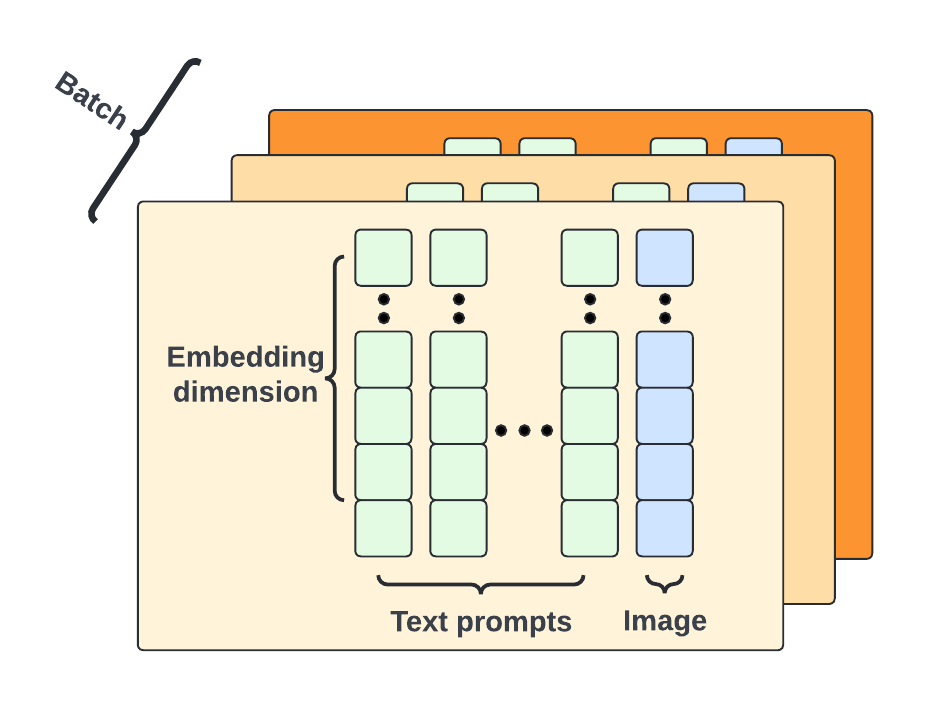}
\caption{Illustration of $\text{Input}_{\text{MHA}}$. Textual representation together with visual one are concatenated into one sequence that is passed thought Masked Multi-Head attention network.}
\label{fig:mha_adapter}
\end{figure}

The Multi-Head attention ($\text{MHA}(\cdot)$) input is created per image sample. 
The text embeddings of prompts $\text{Emb}_{\text{text}}$ and image embeddings of a visual input $\text{Emb}_{\text{image}}$ are concatenated into one sequence:

$$\text{Input}_{\text{MHA}} = \text{Emb}_{\text{text}} \oplus \text{Emb}_{\text{image}} \text{,}$$

where $\text{Input}_{\text{MHA}} \in \mathbb{R}^{(N^{\text{classes}} + 1) \times N^{\text{Batch}} \times E_{N}}$. Here, $N^{\text{classes}} + 1$ represents the summation of the number of classes (textual embeddings) and one image embedding (Figure \ref{fig:mha_adapter}) and $E_{N}$ is dimension of the embeddings.

\paragraph{Masked Multi-Head attention.}
After dimension downsampling, we aggregate latent representations by employing Multi-Head attention layers.
To distinguish the text and image inputs of the Multi-Head attention layer, we use attention masking to stir information interaction within each modality.
We use $0$ for positive interactions and $-\infty$ for the negative ones. Therefore, each textual embedding is adjusted only by the corresponding image information, while partially ignoring information in other textual embeddings. The same is true for image embeddings.
Given $P$ prompts and $I$ images, the total number of elements $T = P + I$:

$$
\mathbf{\text{Mask}}_{i,j} = 
\begin{cases} 
0 & \text{if } (0 \leq i < P \text{ and } P \leq j < T) \\
0 & \text{if } (P \leq i < T \text{ and } 0 \leq j < P) , \\
-\infty & \text{otherwise}
\end{cases}
$$

where $i$ indexes the rows (ranging from $0$ to $T-1$) and $j$ indexes the columns (ranging from $0$ to $T-1$).

After the mask is computed, it is applied to the Multi-Head attention network via sum operation:

\begin{align*}
\quad & \text{Masked Attention}(Q,K,V) = \\  
\quad & \text{softmax}\left( \frac{QK^{T}}{\sqrt{d_{k}}} + \text{Mask}_{i,j} \right) V,
\end{align*}

where $Q$ is the matrix of queries, $K$ is the matrix of keys, $V$ is the matrix of values, $d_{k}$ is the dimensionality of the keys.

\paragraph{Multi-Modal Adapter.}
Afterwards, the output of MHA is passed through two linear upsampling layers $U_{1}(\cdot)$, $U_{2}(\cdot)$ with GELU activation \cite{hendrycks2016gaussian} in between them, where $W_{U1} \in \mathbb{R}^{\text{C}_{\text{Emb} / 4} \times \text{C}_{\text{Emb}} / 16}$ and $W_{U2} \in \mathbb{R}^{\text{C}_{\text{Emb}} / 16 \times \text{C}_{\text{Emb}}}$. 

Overall, we finalize the Multi-Modal Adapter ($\text{Adapter}_{\text{MM}}$) as:

\begin{align*}
\quad & X_{1} = D(\text{Input}_{\text{MHA}}) \\
\quad & X_{2} = \text{MHA}(X_{1}) \\
\quad & X_{3} = U_{1}(X_{2}) \\
\quad & X_{4} = \text{GELU}(X_{3}) \\
\quad & \text{Adapter}_{\text{MM}} = U_{2}(X_{4})
\end{align*}

The output of $\text{Adapter}_{\text{MM}}$ is then split into adapted textual embeddings and adapted visual embedding $(\text{MM Emb}_{\text{text}} ; \text{MM Emb}_{\text{image}})$, and normalized through the $\text{Emb}$ dimension. Lastly, the normalized embeddings $\text{Emb}_{\text{text}}$ and $\text{Emb}_{\text{image}}$ are summed with the outputs of MM using $\lambda$ as a ratio:

\begin{align*}
\lambda \cdot \text{Emb}_{\text{text}} &+ (1 - \lambda) \cdot \text{MM Emb}_{\text{text}},  \\
\lambda \cdot \text{Emb}_{\text{image}} &+ (1 - \lambda) \cdot \text{MM Emb}_{\text{image}}.  
\end{align*}

For our experiments, we set $\lambda$ to $0.2$ for both text and image values, following the CLIP-Adapter approach.

In the end, logits are computed by calculating the cosine similarity between text and image embeddings. We optimize trainable weights $\theta$ using cross-entropy loss:
\[
L(\theta) = -\frac{1}{N}\sum_{n=1}^{N} \sum_{i=1}^{K} y_i^{(n)} \log \hat{y}_i^{(n)},
\]
where \(N\) is the number of each batch training examples; \(y_i^{(n)} = 1\) if \(i\) equals the ground-truth category label \(\hat{i}\), otherwise \(y_i^{(n)} = 0\); \(\hat{y}_i = p_i\) is the predicted probability for class \(i\).

%% file: sections/4_experiments_results.tex
\section{Experiments and Results}
\label{sec:experiments_results}

We construct our main experimentation pipeline employing eleven different visual classification datasets. We primarily focus on few-shot training settings, where the number of samples per class is sixteen. To evaluate the model's ability to perform zero-shot classification after fine-tuning using our adapter, we additionally test how the model performs on unseen classes by splitting the number of classes into two parts: first $N_{\text{classes}} / 2$ classes, denoted as ''base'' (the first half) and ''new'' (the second half). For few-shot experiments, both the training and evaluation sets contain an identical number of samples ($N_{\text{classes}} \cdot N_{\text{samples}}$). Where train and test splits were not provided by the datasets themselves, we used the splits employed in the work of \cite{zhou2022learning}.

Primarily, we train the model on ''base'' classes and evaluate it on ''base'' and ''new'' subsamples of the test set.

We train our model using the Adam optimizer \cite{kingma2014adam}, with a learning rate of $0.005$, a batch size of $256$, and momentum of $0.5$. We employ early stopping with a patience of ten epochs. Our experiments are conducted using an NVIDIA T4 GPU.

We use a Multi-Head Attention network with 4 heads. With the employment of the downsampler mentioned before, we have $107,712$ trainable parameters.

We use the following datasets in our experiments:
Cifar10 \cite{krizhevsky2009learning}, Caltech101 \cite{li2022caltech}, Oxford Pets \cite{parkhi2012cats}, Oxford Flowers \cite{nilsback2008automated}, Food101 \cite{bossard2014food}, Stanford Cars \cite{krause20133d}, FGVC-Aircarft \cite{maji13fine-grained}, DTD \cite{xu2018learning}, SUN397 \cite{xiao2010sun}, Imagenet-V1 \cite{deng2009imagenet}, UCF101 \cite{soomro2012ucf101} and EuroSat \cite{helber2019eurosat}.

\subsection{Few-shot Experiment Results}

\begin{table*}[hbt!]
\setlength\extrarowheight{2pt}
\setlength{\tabcolsep}{3pt}
\centering
\caption{\label{tab:few-shot-results}
Results of few-shot (16 samples per class) training. We evaluate our model, Multi-Modal Adapter, on eleven different datasets and compare it with the results reported by \cite{wang2023improving}. The average accuracy of the base and new classes is represented by the terms \textbf{Base} and
\textbf{New}, respectively, while their harmonic mean is denoted as \textbf{H}. We additionally compute model average performance across all datasets. For the average computation, we also present percentage difference between ''new'' and ''base'' classes accuracies. The best results are presented in bold.}
\resizebox{\textwidth}{!}{\begin{tabular}{l|cccc|ccc|ccc|ccc|ccc|ccc}
\hline
 &
  \multicolumn{4}{c|}{\textbf{Average}} &
  \multicolumn{3}{c|}{\textbf{Caltech101}} &
  \multicolumn{3}{c|}{\textbf{OxfordPets}} &
  \multicolumn{3}{c|}{\textbf{StanfordCars}} &
  \multicolumn{3}{c|}{\textbf{SUN397}} &
  \multicolumn{3}{c}{\textbf{EuroSAT}} \\
 &
  \textbf{Base} &
  \textbf{New} &
  \textbf{H} &
  \textbf{Diff \%} &
  \textbf{Base} &
  \textbf{New} &
  \textbf{H} &
  \textbf{Base} &
  \textbf{New} &
  \textbf{H} &
  \textbf{Base} &
  \textbf{New} &
  \textbf{H} &
  \textbf{Base} &
  \textbf{New} &
  \textbf{H} &
  \textbf{Base} &
  \textbf{New} &
  \textbf{H} \\ \hline
CLIP &
  69.34 &
  \textbf{74.22} &
  71.78 &
  +7 &
  96.84 &
  94 &
  95.4 &
  91.17 &
  97.26 &
  94.12 &
  63.37 &
  \textbf{74.89} &
  69.65 &
  69.36 &
  75.35 &
  72.23 &
  56.48 &
  64.05 &
  60.03 \\
CoOp &
  82.69 &
  63.22 &
  72.955 &
  -31 &
  98 &
  89.81 &
  93.73 &
  93.67 &
  95.29 &
  94.47 &
  78.12 &
  60.4 &
  68.13 &
  \textbf{80.6} &
  65.89 &
  72.51 &
  92.19 &
  54.74 &
  68.69 \\
CoCoOp &
  80.47 &
  71.69 &
  76.08 &
  -12 &
  97.96 &
  93.81 &
  95.84 &
  95.2 &
  97.69 &
  96.43 &
  70.49 &
  73.59 &
  72.01 &
  79.74 &
  76.86 &
  \textbf{78.27} &
  87.49 &
  60.04 &
  71.21 \\
ProDA &
  81.56 &
  72.3 &
  \textbf{76.93} &
  -13 &
  \textbf{98.27} &
  93.23 &
  95.68 &
  \textbf{95.43} &
  \textbf{97.83} &
  \textbf{96.62} &
  74.7 &
  71.2 &
  \textbf{72.91} &
  78.67 &
  76.93 &
  77.79 &
  83.9 &
  \textbf{66} &
  \textbf{73.88} \\
CLIP-Adapter &
  \textbf{83.05} &
  65.2 &
  74.125 &
  -27 &
  98.13 &
  92.19 &
  95.39 &
  91.55 &
  90.1 &
  90.82 &
  \textbf{79.16} &
  59.49 &
  67.93 &
  79.44 &
  66.81 &
  72.58 &
  \textbf{93.45} &
  54.41 &
  68.78 \\
Multi-Modal Adapter (ours) &
  78.37 &
  73.24 &
  75.805 &
  -7 &
  97.8 &
  \textbf{94.1} &
  \textbf{95.93} &
  94.5 &
  97.03 &
  95.8 &
  67.24 &
  70.66 &
  68.95 &
  75.59 &
  \textbf{77.28} &
  76.43 &
  82.14 &
  64.6 &
  73.73 \\ \hline
 \end{tabular}}
 \resizebox{\textwidth}{!}{\begin{tabular}{l|ccc|ccc|ccc|ccc|ccc|ccc}
\hline
&
  \multicolumn{3}{c|}{\textbf{Imagenet}} &
  \multicolumn{3}{c|}{\textbf{Flowers102}} &
  \multicolumn{3}{c|}{\textbf{Food101}} &
  \multicolumn{3}{c|}{\textbf{FGVCAircraft}} &
  \multicolumn{3}{c|}{\textbf{DTD}} &
  \multicolumn{3}{c}{\textbf{UCF101}} \\
 &
  \textbf{Base} &
  \textbf{New} &
  \textbf{H} &
  \textbf{Base} &
  \textbf{New} &
  \textbf{H} &
  \textbf{Base} &
  \textbf{New} &
  \textbf{H} &
  \textbf{Base} &
  \textbf{New} &
  \textbf{H} &
  \textbf{Base} &
  \textbf{New} &
  \textbf{H} &
  \textbf{Base} &
  \textbf{New} &
  \textbf{H} \\ \hline
CLIP &
  72.43 &
  68.14 &
  70.22 &
  72.08 &
  77.8 &
  74.83 &
  90.1 &
  91.22 &
  90.66 &
  27.19 &
  \textbf{36.29} &
  31.09 &
  53.24 &
  \textbf{59.9} &
  56.37 &
  70.53 &
  \textbf{77.5} &
  73.85 \\
CoOp &
  \textbf{76.47} &
  67.88 &
  \textbf{71.92} &
  97.6 &
  59.67 &
  74.06 &
  88.33 &
  82.26 &
  85.19 &
  40.44 &
  22.3 &
  28.75 &
  79.44 &
  41.18 &
  54.24 &
  84.69 &
  56.05 &
  67.46 \\
CoCoOp &
  75.98 &
  \textbf{70.43} &
  73.1 &
  94.87 &
  71.75 &
  \textbf{81.71} &
  90.7 &
  91.29 &
  90.99 &
  33.41 &
  23.71 &
  27.74 &
  77.01 &
  56 &
  64.85 &
  82.33 &
  73.45 &
  77.64 \\
ProDA &
  75.4 &
  70.23 &
  72.72 &
  97.7 &
  68.68 &
  80.66 &
  90.3 &
  88.57 &
  89.43 &
  36.9 &
  34.13 &
  \textbf{35.46} &
  80.67 &
  56.48 &
  \textbf{66.44} &
  85.23 &
  71.97 &
  78.04 \\
CLIP-Adapter &
  75.74 &
  68.21 &
  71.78 &
  \textbf{98.29} &
  64.68 &
  78.02 &
  88.24 &
  88.33 &
  88.29 &
  \textbf{42.14} &
  25.67 &
  31.91 &
  \textbf{81.94} &
  39.49 &
  53.3 &
  \textbf{85.42} &
  67.77 &
  75.58 \\
Multi-Modal Adapter (ours) &
  73.23 &
  69.52 &
  71.38 &
  91.26 &
  73.4 &
  76.21 &
  \textbf{92.57} &
  \textbf{93.3} &
  \textbf{92.93} &
  34.15 &
  35.27 &
  33.51 &
  71.84 &
  54.58 &
  63.04 &
  81.79 &
  75.95 &
  \textbf{78.24} \\ \hline
\end{tabular}}

\end{table*}

We present the few-shot experiment results for all models across all datasets in Table \ref{tab:few-shot-results}, highlighting the difference between the accuracy of ''base'' and ''new'' class subsets. 
Our findings indicate that while Multi-Modal Adapter achieves the best performance on some datasets (Caltech101, Food101), it underperforms on others (Oxford Flowers, FGVC-Aircraft). Moreover, while improving over CLIP Base results, Multi-Modal Adapter is outperformed by existing approaches.

However, when examining the accuracy difference between ''new'' and ''base'' class subsamples, we observe our approach maintains more consistent performance across data it was trained on and unseen data. Moreover, Multi-Modal Adapter decreases accuracy on new classes the least, while significantly improvement the accuracy over both base classes. It is possible that this trade-off is achieved by Multi-Modal Adapter learning less and thus forgetting less.

\begin{figure}[hbt!]
\centering
\includegraphics[width=0.7\linewidth]{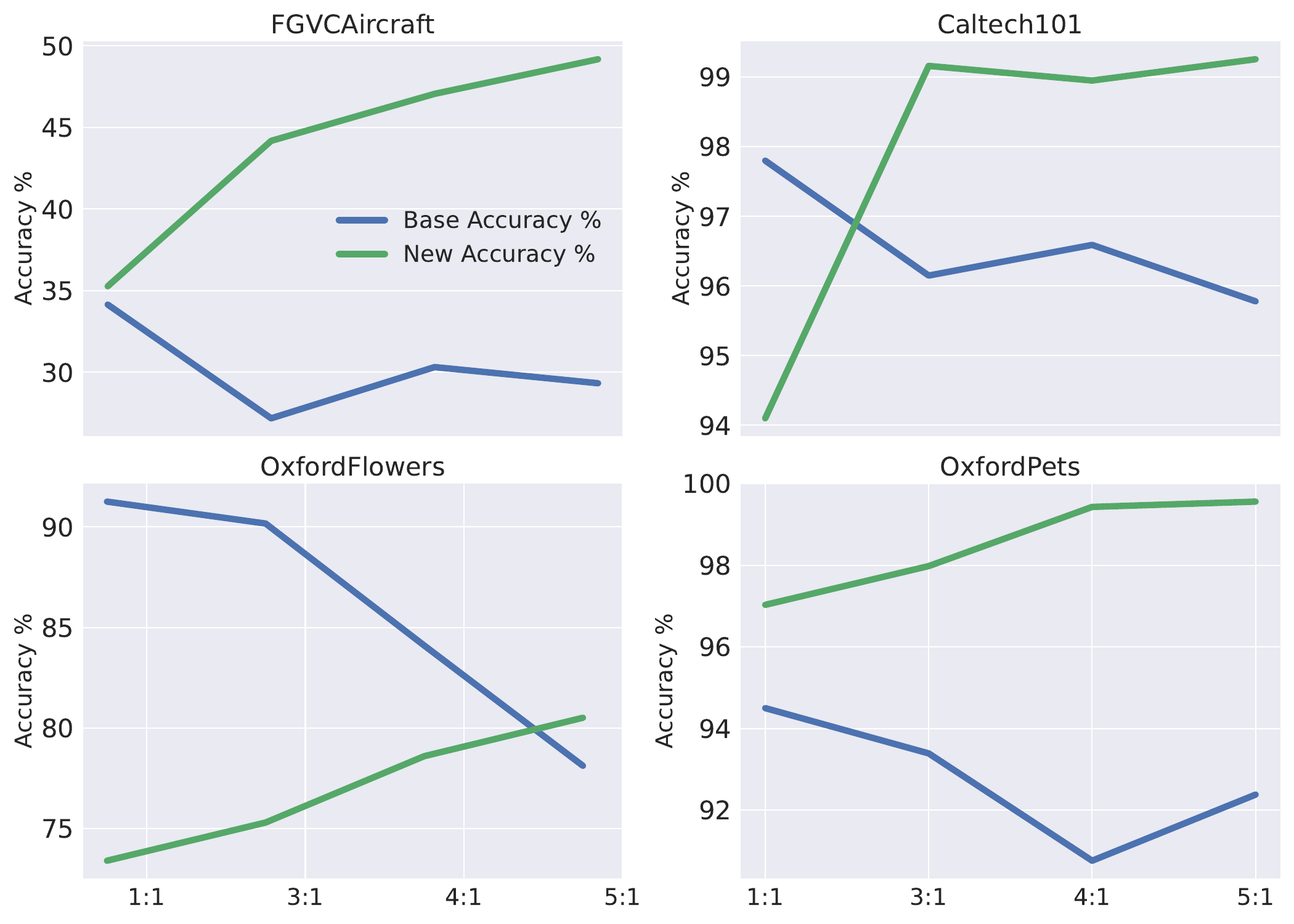}
\caption{Ablation studies on the impact of class subsampling. The graph below shows test accuracy results for both "base" and "new" class splits, evaluated under different class share proportions. \vspace{-10pt}}
\label{fig:split_ab}
\end{figure}

As our approach adapts both visual and textual embeddings, we believe this allows our model to better regularize and reduce overfitting. This is particularly evident in Figure \ref{fig:split_ab}, where we see that the model does not overfit even as we increase the share of classes we train versus test on. 

\subsection{Inspecting Role of Text Adaptation}
To examine the importance of text adaptation in our Multi-Modal Adapter, we conducted experiments on four different visual datasets. We compared the accuracy differences between adapters with and without text adaptation. The results are presented in Table \ref{tab:text_exp}. As indicated by these results, the adapter with text adaptation consistently outperforms the one without text adaptation. In tasks where even small differences in accuracy are crucial, a 1\% improvement is significant. The positive effect of text adaptation is especially present within Oxford Flowers dataset, where without it, model fails to significantly increase accuracy for the ''base'' classes.

\begin{table*}
\setlength\extrarowheight{2pt}
\setlength{\tabcolsep}{6pt}
\centering
\caption{\label{tab:text_exp}
Accuracy comparison for ''base'' and ''new'' subsamples between Multi-Modal Adapter with and without text adaptation.}
\resizebox{\textwidth}{!}{\begin{tabular}{l|cccc}
\hline
\multicolumn{1}{c|}{\multirow{2}{*}{Dataset}} & \textbf{Base acc \%}       & \textbf{New acc \%}        & \textbf{Base acc \%}         & \textbf{New acc \%}          \\
\multicolumn{1}{c|}{}                         & \textbf{w text adaptation} & \textbf{w text adaptation} & \textbf{w/o text adaptation} & \textbf{w/o text adaptation} \\ \hline
FGVCAircraft  & 34.15 & 35.27 & 31.58 & 34.61 \\
Caltech101    & 97.8  & 93.9  & 97.58 & 94.1  \\
OxfordFlowers & 91.26 & 73.4 & 75.97 & 75.31 \\
OxfordPets    & 94.5  & 97.03 & 94.38 & 96.34 \\ \hline
\end{tabular}}
\end{table*}

%% file: sections/5_conclusion.tex
\section{Conclusion}
\label{sec:conclusion}

In this work, we propose a Multi-Modal Adapter for adapting vision-based datasets to specific tasks. We jointly process CLIP's visual and textual embeddings through a Multi-Head attention network, utilizing its outputs to separately adapt visual and textual representations. Our studies indicate that while our approach does not excel in all tasks, it outperforms the alternatives in several instances. Notably, our adapter loses the least performance on new classes as compared to CLIP, while significantly improving over base class performance.

We further explore the effect of textual representation adaptation within our Multi-Modal Adapter framework. Our findings demonstrate that the combination of both textual and visual representation adaptation achieves superior performance compared to adapting only visual representations.

We acknowledge that extensive configuration settings remain unexplored, as our experiments are based on a single Multi-Head attention setting. A more comprehensive hyperparameter search is necessary to fully investigate the capabilities of our approach. Future work should also consider the use of positional encodings for textual and visual features within the sequence of the Multi-Head attention network, alongside attention masking.

Despite some limitations, our findings motivate further exploration of the joint adaptation of both visual and textual representations in zero-shot and few-shot classification settings.

%% file: sections/appendix.tex

\newpage
\newpage
\appendix
\section{Appendix}
\subsection{Additional ablations}
\begin{table*}[hbt!]
\setlength\extrarowheight{2pt}
\setlength{\tabcolsep}{4pt}
\centering
\caption{Ablations of adapter configurations to compare a) use of multi-head attention (MHA) vs transformer-encoder block and b) up-/downsampling using a linear layer vs MLP. \textbf{Base} represents baseline accuracy, \textbf{New} represents accuracy on new samples, and \textbf{All} represents overall test accuracy.}
\resizebox{\textwidth}{!}{\begin{tabular}{l|ccc|ccc|ccc}
\hline
 & \multicolumn{3}{c|}{\textbf{Caltech101}} & \multicolumn{3}{c|}{\textbf{Oxford Pets}} & \multicolumn{3}{c}{\textbf{CIFAR-10}} \\
 & \textbf{Base} & \textbf{New} & \textbf{All} & \textbf{Base} & \textbf{New} & \textbf{All} & \textbf{Base} & \textbf{New} & \textbf{All} \\ \hline
CLIP Baseline & 97.159 & 94.105 & 93.306 & 90.909 & \textbf{97.032} & 89.098 & 96.08 & 94.24 & 90.15 \\
MHA adapter, linear up-/downsampling & \textbf{97.547} & 93.777 & 94.199 & 93.883 & 96.025 & 88.771 & 95.82 & 94.42 & 89.74 \\
MHA adapter, MLP up-/downsampling & 97.482 & 93.886 & 94.037 & \textbf{94.444} & 96.184 & 88.771 & 95.9 & \textbf{95.22} & \textbf{90.37} \\
Transformer adapter, linear up-/downsampling & 97.482 & \textbf{94.541} & \textbf{94.361} & 93.883 & 95.76 & 89.261 & \textbf{96.12} & 94.44 & 89.86 \\
Transformer adapter, MLP up-/downsampling  & 97.353 & 93.996 & 94.158 & 94.332 & \textbf{97.032} & \textbf{89.507} & 93.3 & 92.68 & 87.61 \\
\hline
\end{tabular}}

\vspace{0.4em}

\resizebox{\textwidth}{!}{\begin{tabular}{l|ccc|ccc|ccc}
\hline
 & \multicolumn{3}{c|}{\textbf{SUN397}} 
 & \multicolumn{3}{c|}{\textbf{FGVC Aircraft}} & \multicolumn{3}{c}{\textbf{DTD}} \\
 & \textbf{Base} & \textbf{New} & \textbf{All} 
 & \textbf{Base} & \textbf{New} & \textbf{All} 
 & \textbf{Base} & \textbf{New} & \textbf{All} \\ \hline
CLIP Baseline
& 69.586 & 75.05 & 62.302 
& 27.671 & 35.753 & 24.902 
& 52.935 & 59.375 & 44.043 \\
MHA adapter, linear up-/downsampling
& \textbf{73.727} & 77.296 & 66.222 
& \textbf{32.053} & 35.153 & \textbf{25.743}
& \textbf{57.935} & 59.896 & \textbf{46.117} \\
MHA adapter, MLP up-/downsampling 
& 73.717 & \textbf{77.317} & \textbf{66.237}
& 31.092 & 35.393 & 25.053 
& 57.5 & \textbf{60.833} & 45.479 \\
Transformer adapter, linear up-/downsampling
& 73.646 & \textbf{77.317} & 66.116 
& 29.952 & 36.053 & 25.413 
& 55.543 & 59.375 & 43.989 \\
Transformer adapter, MLP up-/downsampling  
& 73.626 & 77.226 & 66.131 
& 30.792 & \textbf{36.893} & \textbf{25.743}
& 56.848 & 59.479 & 45.053 \\
\hline
\end{tabular}}
\end{table*}

To validate our design choices for the Multi-Modal Adapter, we conduct ablations, comparing the use of multi-head attention (MHA) versus the transformer-encoder block, and up-/downsampling using a linear layer versus an MLP. Since we observed no significant differences in performance between these alternatives, we opted to use multi-head attention. For up-/downsampling, we implemented linear layers for downsampling and a 2-layer MLP for upsampling, paralleling the approach used in the CLIP Adapter architecture.

\subsection{Noisy training data}

\begin{table*}[hbt!]
\setlength\extrarowheight{2pt}
\setlength{\tabcolsep}{6pt}
\centering
\caption{Accuracy on the original test set after training on a noisy training set. \textbf{Base} represents baseline accuracy, \textbf{New} represents accuracy on new samples, and \textbf{All} represents overall test accuracy.}
\resizebox{\textwidth}{!}{\begin{tabular}{l|ccc|ccc|ccc}
\hline
 & \multicolumn{3}{c|}{\textbf{Caltech101}} & \multicolumn{3}{c|}{\textbf{Oxford Pets}} & \multicolumn{3}{c}{\textbf{CIFAR-10}} \\
 & \textbf{Base} & \textbf{New} & \textbf{All} & \textbf{Base} & \textbf{New} & \textbf{All} & \textbf{Base} & \textbf{New} & \textbf{All} \\ \hline
CLIP Baseline & 97.159 & 94.105 & 93.306 & 90.909 & \textbf{97.032} & 89.098 & 96.08 & 94.24 & 90.15 \\
CLIP Adapter & 97.547 & \textbf{94.432} & 93.712 & \textbf{94.444} & 96.714 & \textbf{90.951} & 95.08 & 93.34 & 88.79 \\
Multi-modal Adapter (ours) & \textbf{97.74} & 94.323 & \textbf{94.402} & 94.108 & \textbf{97.032} & 90.461 & \textbf{96.18} & \textbf{95.44} & \textbf{90.8} \\
\hline
\end{tabular}}

\vspace{0.4em}

\resizebox{\textwidth}{!}{\begin{tabular}{l|ccc|ccc|ccc}
\hline
 & \multicolumn{3}{c|}{\textbf{SUN397}} & \multicolumn{3}{c|}{\textbf{FGVC Aircraft}} & \multicolumn{3}{c}{\textbf{DTD}} \\
 & \textbf{Base} & \textbf{New} & \textbf{All} & \textbf{Base} & \textbf{New} & \textbf{All} & \textbf{Base} & \textbf{New} & \textbf{All} \\ \hline
CLIP Baseline & 69.586 & 75.05 & 62.302 & 27.671 & 35.753 & 24.902 & 52.935 & \textbf{59.375} & 44.043 \\
CLIP Adapter & \textbf{80.364} & 71.005 & 65.023 & \textbf{40.576} & 28.254 & \textbf{26.523} & \textbf{78.913} & 47.5 & 47.394 \\
Multi-modal Adapter (ours) & 75.919 & \textbf{76.925} & \textbf{66.831} & 29.652 & \textbf{36.353} & 25.383 & 68.37 & 57.187 & \textbf{47.713} \\
\hline
\end{tabular}}
\end{table*}

\begin{figure}
\centering
\includegraphics[width=0.9\linewidth]{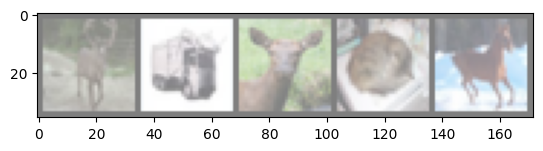}
\includegraphics[width=0.9\linewidth]{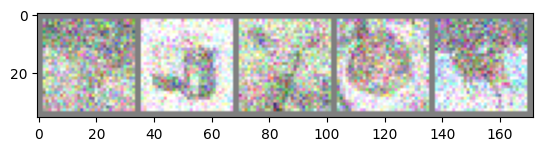}
\caption{Illustration of training set images for CIFAR-10 after applying Gaussian noise.}
\label{fig:noisy_train_set}
\end{figure}

We hypothesized that the Multi-Modal Adapter approach would show robust performance on the original test set when trained on noisy data, due to reduced over-fitting. To validate this hypothesis, we applied Gaussian noise to the training data, as illustrated in Figure \ref{fig:noisy_train_set}. We observed that the Multi-Modal Adapter yields robust performance, outperforming the CLIP Baseline and mostly improving over the CLIP Adapter.

%% file: main.bbl
\begin{thebibliography}{10}
\providecommand{\url}[1]{\texttt{#1}}
\providecommand{\urlprefix}{URL }
\providecommand{\doi}[1]{https://doi.org/#1}

\bibitem{alayrac2022flamingo}
Alayrac, J.B., Donahue, J., Luc, P., Miech, A., Barr, I., Hasson, Y., Lenc, K., Mensch, A., Millican, K., Reynolds, M., et~al.: Flamingo: a visual language model for few-shot learning. In: Advances in neural information processing systems. vol.~35, pp. 23716--23736 (2022)

\bibitem{bao2022vlmo}
Bao, H., Wang, W., Dong, L., Liu, Q., Mohammed, O.K., Aggarwal, K., Som, S., Wei, F.: Vlmo: Unified vision-language pre-training with mixture-of-modality-experts (2022)

\bibitem{bossard2014food}
Bossard, L., Guillaumin, M., Van~Gool, L.: Food-101--mining discriminative components with random forests. In: Computer Vision--ECCV 2014: 13th European Conference, Zurich, Switzerland, September 6-12, 2014, Proceedings, Part VI 13. pp. 446--461. Springer (2014)

\bibitem{deng2009imagenet}
Deng, J., Dong, W., Socher, R., Li, L.J., Li, K., Fei-Fei, L.: Imagenet: A large-scale hierarchical image database. In: 2009 IEEE conference on computer vision and pattern recognition. pp. 248--255. Ieee (2009)

\bibitem{dosovitskiy2020image}
Dosovitskiy, A., Beyer, L., Kolesnikov, A., Weissenborn, D., Zhai, X., Unterthiner, T., Dehghani, M., Minderer, M., Heigold, G., Gelly, S., et~al.: An image is worth 16x16 words: Transformers for image recognition at scale. arXiv preprint arXiv:2010.11929  (2020)

\bibitem{gao2021clipadapter}
Gao, P., Geng, S., Zhang, R., Ma, T., Fang, R., Zhang, Y., Li, H., Qiao, Y.: Clip-adapter: Better vision-language models with feature adapters (2021)

\bibitem{gao2024clip}
Gao, P., Geng, S., Zhang, R., Ma, T., Fang, R., Zhang, Y., Li, H., Qiao, Y.: Clip-adapter: Better vision-language models with feature adapters. International Journal of Computer Vision  \textbf{132}(2),  581--595 (2024)

\bibitem{girdhar2023imagebind}
Girdhar, R., El-Nouby, A., Liu, Z., Singh, M., Alwala, K.V., Joulin, A., Misra, I.: Imagebind: One embedding space to bind them all (2023)

\bibitem{he2015deep}
He, K., Zhang, X., Ren, S., Sun, J.: Deep residual learning for image recognition. corr abs/1512.03385 (2015) (2015)

\bibitem{helber2019eurosat}
Helber, P., Bischke, B., Dengel, A., Borth, D.: Eurosat: A novel dataset and deep learning benchmark for land use and land cover classification. IEEE Journal of Selected Topics in Applied Earth Observations and Remote Sensing  \textbf{12}(7),  2217--2226 (2019)

\bibitem{hendrycks2016gaussian}
Hendrycks, D., Gimpel, K.: Gaussian error linear units (gelus). arXiv preprint arXiv:1606.08415  (2016)

\bibitem{houlsby2019parameterefficient}
Houlsby, N., Giurgiu, A., Jastrzebski, S., Morrone, B., de~Laroussilhe, Q., Gesmundo, A., Attariyan, M., Gelly, S.: Parameter-efficient transfer learning for nlp (2019)

\bibitem{hu2021lora}
Hu, E.J., Shen, Y., Wallis, P., Allen-Zhu, Z., Li, Y., Wang, S., Wang, L., Chen, W.: Lora: Low-rank adaptation of large language models (2021)

\bibitem{jia2021scaling}
Jia, C., Yang, Y., Xia, Y., Chen, Y.T., Parekh, Z., Pham, H., Le, Q.V., Sung, Y., Li, Z., Duerig, T.: Scaling up visual and vision-language representation learning with noisy text supervision (2021)

\bibitem{kingma2014adam}
Kingma, D.P., Ba, J.: Adam: A method for stochastic optimization. arXiv preprint arXiv:1412.6980  (2014)

\bibitem{krause20133d}
Krause, J., Stark, M., Deng, J., Fei-Fei, L.: 3d object representations for fine-grained categorization. In: Proceedings of the IEEE international conference on computer vision workshops. pp. 554--561 (2013)

\bibitem{krizhevsky2009learning}
Krizhevsky, A., Hinton, G., et~al.: Learning multiple layers of features from tiny images  (2009)

\bibitem{lester2021power}
Lester, B., Al-Rfou, R., Constant, N.: The power of scale for parameter-efficient prompt tuning (2021)

\bibitem{li2022caltech}
Li, F.F., Andreeto, M., Ranzato, M., Perona, P.: Caltech 101. CaltechDATA: Pasadena, CA, USA  (2022)

\bibitem{li2021prefixtuning}
Li, X.L., Liang, P.: Prefix-tuning: Optimizing continuous prompts for generation (2021)

\bibitem{liu2024dora}
Liu, S.Y., Wang, C.Y., Yin, H., Molchanov, P., Wang, Y.C.F., Cheng, K.T., Chen, M.H.: Dora: Weight-decomposed low-rank adaptation (2024)

\bibitem{maji13fine-grained}
Maji, S., Kannala, J., Rahtu, E., Blaschko, M., Vedaldi, A.: Fine-grained visual classification of aircraft. Tech. rep. (2013)

\bibitem{nilsback2008automated}
Nilsback, M.E., Zisserman, A.: Automated flower classification over a large number of classes. In: 2008 Sixth Indian conference on computer vision, graphics \& image processing. pp. 722--729. IEEE (2008)

\bibitem{openai2024gpt4}
OpenAI, Achiam, J., Adler, S., Agarwal, S., Ahmad, L., Akkaya, I., Aleman, F.L., Almeida, D., Altenschmidt, J., Altman, S., Anadkat, S., et~al.: Gpt-4 technical report (2024)

\bibitem{parkhi2012cats}
Parkhi, O.M., Vedaldi, A., Zisserman, A., Jawahar, C.: Cats and dogs. In: 2012 IEEE conference on computer vision and pattern recognition. pp. 3498--3505. IEEE (2012)

\bibitem{radford2021learning}
Radford, A., Kim, J.W., Hallacy, C., Ramesh, A., Goh, G., Agarwal, S., Sastry, G., Askell, A., Mishkin, P., Clark, J., Krueger, G., Sutskever, I.: Learning transferable visual models from natural language supervision (2021)

\bibitem{rebuffi2017learning}
Rebuffi, S.A., Bilen, H., Vedaldi, A.: Learning multiple visual domains with residual adapters (2017)

\bibitem{soomro2012ucf101}
Soomro, K., Zamir, A.R., Shah, M.: Ucf101: A dataset of 101 human actions classes from videos in the wild. arXiv preprint arXiv:1212.0402  (2012)

\bibitem{geminiteam2024gemini}
Team, G., Anil, R., Borgeaud, S., Wu, Y., Alayrac, J.B., Yu, J., Soricut, R., Schalkwyk, J., Dai, A.M., Hauth, A., et~al.: Gemini: a family of highly capable multimodal models. arXiv preprint arXiv:2312.11805  (2023)

\bibitem{tsimpoukelli2021multimodal}
Tsimpoukelli, M., Menick, J., Cabi, S., Eslami, S.M.A., Vinyals, O., Hill, F.: Multimodal few-shot learning with frozen language models (2021)

\bibitem{vaswani2017attention}
Vaswani, A., Shazeer, N., Parmar, N., Uszkoreit, J., Jones, L., Gomez, A.N., Kaiser, {\L}., Polosukhin, I.: Attention is all you need. Advances in neural information processing systems  \textbf{30} (2017)

\bibitem{wang2023improving}
Wang, Z., Liang, J., He, R., Xu, N., Wang, Z., Tan, T.: Improving zero-shot generalization for clip with synthesized prompts. In: Proceedings of the IEEE/CVF International Conference on Computer Vision. pp. 3032--3042 (2023)

\bibitem{xiao2010sun}
Xiao, J., Hays, J., Ehinger, K.A., Oliva, A., Torralba, A.: Sun database: Large-scale scene recognition from abbey to zoo. In: 2010 IEEE computer society conference on computer vision and pattern recognition. pp. 3485--3492. IEEE (2010)

\bibitem{xiao2024any}
Xiao, Z., Shen, J., Derakhshani, M.M., Liao, S., Snoek, C.G.M.: Any-shift prompting for generalization over distributions. In: CVPR (2024)

\bibitem{xu2018learning}
Xu, Z., Wilber, M., Fang, C., Hertzmann, A., Jin, H.: Learning from multi-domain artistic images for arbitrary style transfer. arXiv preprint arXiv:1805.09987  (2018)

\bibitem{yang2024crossmodal}
Yang, J., Li, Z., Xie, S., Zhu, W., Yu, W., Li, S.: Cross-modal adapter: Parameter-efficient transfer learning approach for vision-language models (2024)

\bibitem{zhang2021tipadapter}
Zhang, R., Fang, R., Zhang, W., Gao, P., Li, K., Dai, J., Qiao, Y., Li, H.: Tip-adapter: Training-free clip-adapter for better vision-language modeling (2021)

\bibitem{zhou2022conditional}
Zhou, K., Yang, J., Loy, C.C., Liu, Z.: Conditional prompt learning for vision-language models. In: Proceedings of the IEEE/CVF conference on computer vision and pattern recognition. pp. 16816--16825 (2022)

\bibitem{Zhou_2022}
Zhou, K., Yang, J., Loy, C.C., Liu, Z.: Learning to prompt for vision-language models. International Journal of Computer Vision  \textbf{130}(9),  2337–2348 (Jul 2022). \doi{10.1007/s11263-022-01653-1}, \url{http://dx.doi.org/10.1007/s11263-022-01653-1}

\bibitem{zhou2022learning}
Zhou, K., Yang, J., Loy, C.C., Liu, Z.: Learning to prompt for vision-language models. International Journal of Computer Vision  \textbf{130}(9),  2337--2348 (2022)

\end{thebibliography}
